\documentclass[letterpaper, 10 pt, journal, twoside]{IEEEtran}
\IEEEoverridecommandlockouts
\usepackage{amsmath,amssymb,amsfonts}
\usepackage{algorithmic}
\usepackage{graphicx}
\usepackage{textcomp}
\usepackage{xcolor}
\usepackage{amsmath}
\usepackage{amssymb}
\usepackage{booktabs}
\usepackage{array}
\usepackage{bm}
\usepackage{comment}
\usepackage{subfig}
\usepackage{xspace}
\usepackage{multirow}
\usepackage{soul}
\usepackage{pbalance}
\usepackage[ruled,linesnumbered]{algorithm2e}
\SetKwInput{KwData}{Input}
\SetKwInput{KwResult}{Output}
\newcommand{\rebuttal}[1]{\textcolor{black}{#1}}

\newcommand{\etal}{\textit{et~al.}\xspace}
\usepackage[backend=biber,style=numeric-comp,sorting=none,maxbibnames=6]{biblatex}
\def\BibTeX{{\rm B\kern-.05em{\sc i\kern-.025em b}\kern-.08em
    T\kern-.1667em\lower.7ex\hbox{E}\kern-.125emX}}
\newcommand{\algoName}{\texttt{NeRF-Loc}\xspace} %option 2
\newcommand{\benchmarkName}{\texttt{NeRFLocBench}\xspace} %option 2
\bibliography{conference}

\begin{document}

\title{
\algoName: Transformer-Based Object Localization Within Neural Radiance Fields
}

\author{
Jiankai Sun$^{1,*}$, Yan Xu$^{2,*}$, Mingyu Ding$^{3}$, Hongwei Yi$^{4}$, Chen Wang$^{1}$, \\ 
Jingdong Wang$^{5}$ \textit{Fellow, IEEE}, Liangjun Zhang$^{6}$, Mac Schwager$^{1}$~\textit{Member, IEEE} 
\thanks{Manuscript received: March 10, 2023; Revised: May 28, 2023; Accepted: June 19, 2023.}%Use only for final RAL version
\thanks{This paper was recommended for publication by Editor Aleksandra Faust upon evaluation of the Associate Editor and Reviewers' comments.}
\thanks{$^{1} $Jiankai Sun, Chen Wang, and Mac Schwager are with the School of Engineering, Stanford University, Stanford, CA 94305, USA (email: \texttt{jksun@stanford.edu}).}%
\thanks{$^{2} $Yan Xu is with the Chinese University of Hong Kong, Hong Kong (email: \texttt{yanxu@link.cuhk.edu.hk}).}%
\thanks{$^{3} $Mingyu Ding is with UC Berkeley, USA.} 
\thanks{$^{4} $Hongwei Yi is with Max Planck Institute for Intelligent Systems, T\"ubingen, Germany.}%
\thanks{$^{5} $Jingdong Wang is with Baidu Inc.}%
\thanks{$^{6} $Liangjun Zhang is with Robotics and Autonomous Driving Lab of Baidu Research, Sunnyvale, CA 94089 USA.}%
\thanks{
$^* $denotes equal contribution. 
}
% \thanks{
% $^\dagger $work done as an intern at Baidu Research.
% }
% single correspondence email, since other emails may not last long
\thanks{Digital Object Identifier (DOI): see the top of this page.}
}

\markboth{IEEE Robotics and Automation Letters. Preprint Version. Accepted June, 2023}{Sun \MakeLowercase{\textit{et al.}}: \algoName: Transformer-Based Object Localization Within Neural Radiance Fields}
\maketitle
\begin{abstract}
Neural Radiance Fields (NeRFs) have become a widely-applied scene representation technique in recent years, showing advantages for robot navigation and manipulation tasks.
 To further advance the utility of NeRFs for robotics, we propose a transformer-based framework, \algoName, to extract 3D bounding boxes of objects in NeRF scenes. \algoName takes a pre-trained NeRF model and camera view as input and produces labeled, oriented 3D bounding boxes of objects as output.  Using current NeRF training tools, a robot can train a NeRF environment model in real-time and, using our algorithm, identify 3D bounding boxes of objects of interest within the NeRF for downstream navigation or manipulation tasks.
Concretely, we design a pair of paralleled transformer encoder branches, namely the coarse stream and the fine stream, to encode both the context and details of target objects. The encoded features are then fused together with attention layers to alleviate ambiguities for accurate object localization. 
We have compared our method with conventional RGB(-D) based methods that take rendered RGB images and depths from NeRFs as inputs. Our method is better than the baselines. 
\end{abstract}


\section{Introduction}

\begin{figure}
    \centering
    \subfloat[Architecture]{\includegraphics[width=\linewidth]{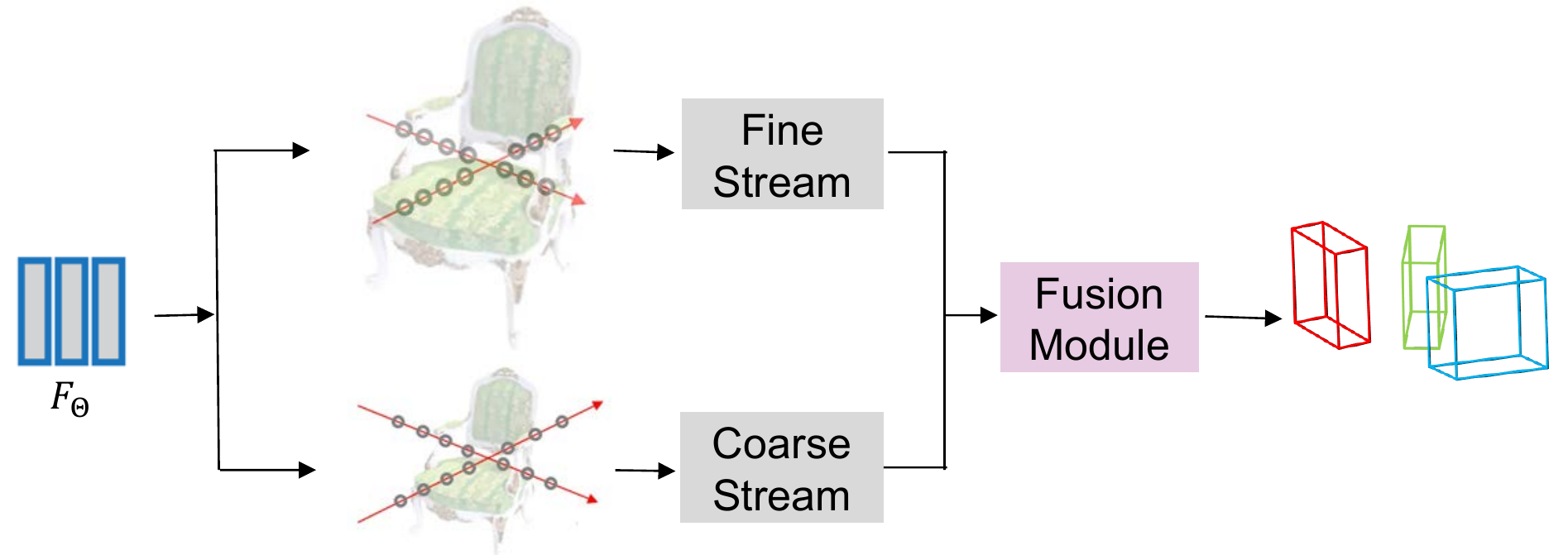}\label{subfig:arch}}
    \vspace{0.1cm}
    \subfloat[$\times 1.5$]{\includegraphics[width=0.24\linewidth]{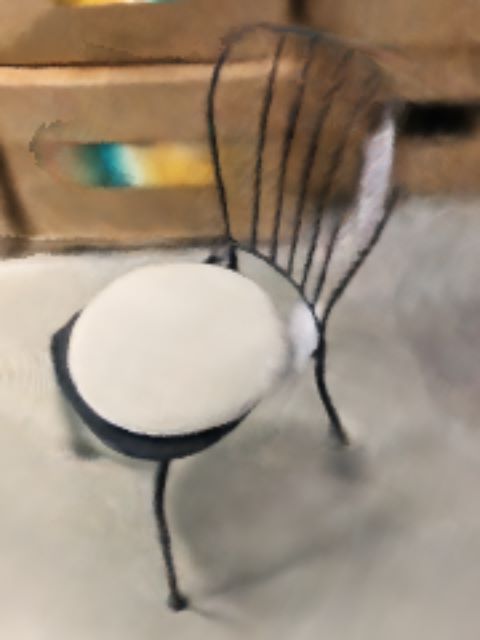}\label{subfig:divide_0.7}}
    \hspace{0.001cm}
    \subfloat[$\times 1$]{\includegraphics[width=0.24\linewidth]{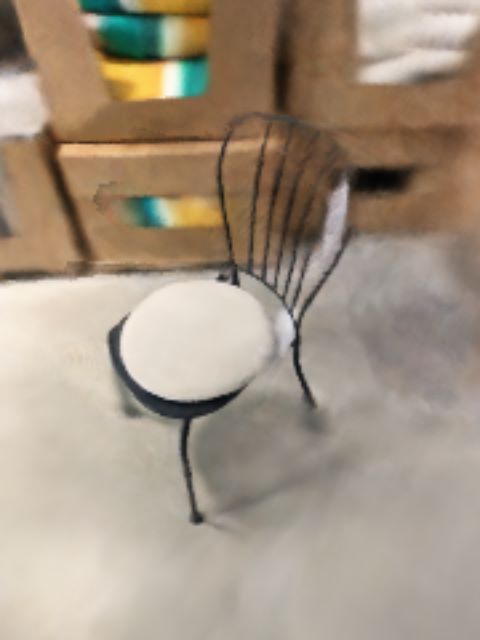}\label{subfig:divide_1.0}}
    \hspace{0.001cm}
    \subfloat[$\times 0.67$]{\includegraphics[width=0.24\linewidth]{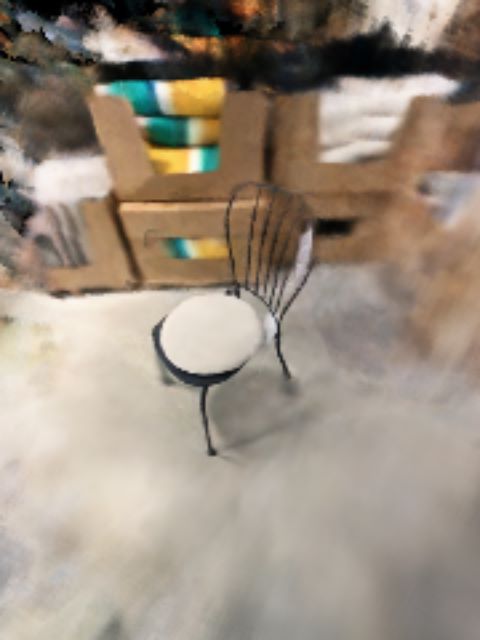}\label{subfig:divide_1.5}}
    \hspace{0.001cm}
    \subfloat[$\times 0.5$]{\includegraphics[width=0.24\linewidth]{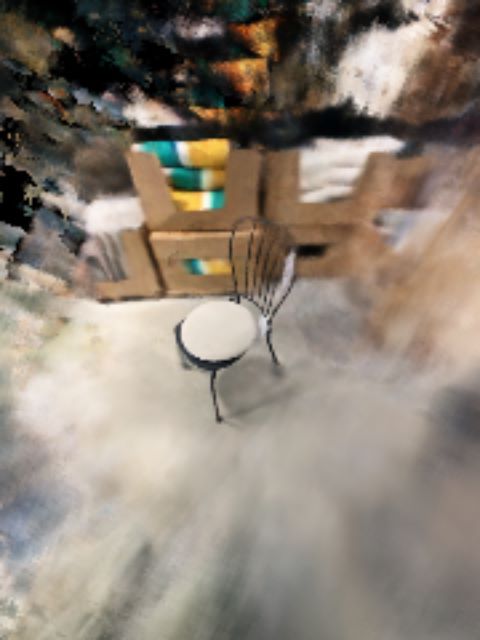}\label{subfig:divide_2.0}}
    \caption{(a) \algoName networks process information at multiple scales. The coarse stream learns to sample the most informative positions in the overview image, while the fine stream processes the near-field image to extract fine-grained information. These two streams are connected by an attention fusion module, after which 3D object localizations are predicted. (b)-(e) shows images rendered by NeRF from fine to coarse scales. We can see that they contain different amounts of information.}
    \label{fig:teaser}
    \vspace{-12pt}
\end{figure}
\IEEEPARstart{W}{hen} a robot enters a novel environment, it needs to first perceive the surrounding objects and understand their spatial relationships, so that it can navigate toward objects of interest or avoid objects that may present a threat.  Similarly, for manipulation tasks, a robot must first detect the object it intends to manipulate, and determine its pose relative to the robot.  
Accurate object localization in 3D space thus becomes a fundamental problem in robotics.  
The choice of object localization strategy depends on the underlying map representation~\cite{singh2020moca,sun2022plate}.
Compared with representing the scene with a set of 2D images, 3D maps maintain 3D topology and richer geometric information. Moving further towards high-fidelity 3D environment representations, in this paper, we propose a method to detect 3D object bounding boxes directly within a NeRF environment model.

\begin{figure*}[htbp]
    \centering
    \includegraphics[width=0.9\linewidth]{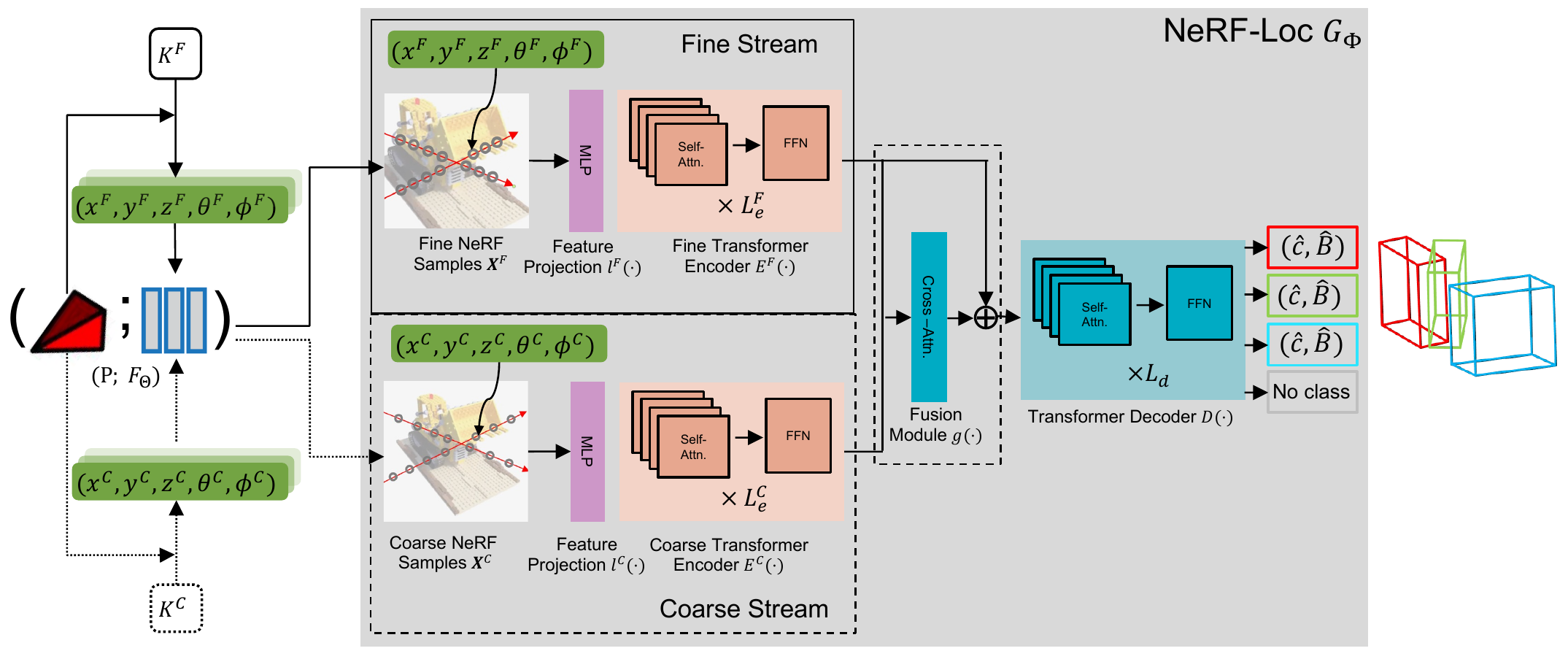}
    \caption{\textbf{Framework.} 
    Given the observation camera pose $\mathbf{P}\in SE(3)$, 
    we sample a set of field values along view rays 
    ($(x,y,z)$'s) and  aim to localize the objects based on these samples.
    Our framework contains a fine stream and a coarse stream. The two streams share similar transformer architectures but are dedicated to processing NeRF samples of different fields of view (controlled by intrinsics $K^C$ and $K^F$). 
    The samples from two streams are encoded as high-dimensional embeddings separately before being fused together by the  cross-attention-based Fusion Module. 
    The fused features are decoded by the Transformer Decoder. Finally, the 3D bounding boxes, and the object categories are predicted by MLP heads. 
    }
    \label{fig:framework}
    \vspace{-10pt}
\end{figure*}
Point cloud or voxel-based map representations have significant limitations. 
Recently, researchers have identified this issue and proposed to represent the environment with Neural Radiance Fields (NeRFs)~\cite{mildenhall2020nerf}, and successfully deploy it in real-time robot SLAM~\cite{sucar2021imap,zhu2022nice} systems.
With modern NeRF training tools, a robot can build a high-fidelity NeRF model of its environment, and use this as a map representation for downstream navigation and manipulation tasks.  NeRF is a continuous scene representation where the 3D scene is compressed in feed-forward neural networks that memorize the 3D spatial density and radiance fields. Compared with classic point cloud and voxel representations, NeRF is more compact and contains more photo-realistic elements.   
\rebuttal{Other motivations for utilizing NeRF in the field of robotics include the potential benefits it offers in terms of enabling end-to-end robot learning, acting as a robust simulation platform, and enhancing robot performance by training in simulated NeRF environments.}
Nevertheless, conventional object localization methods cannot be applied directly to NeRFs, and object localization in NeRFs has not been well-studied in the literature. Object detection is crucial for specifying objectives for robotic motion planning~\cite{adamkiewicz2022vision} and manipulation~\cite{IchnowskiAvigal2021DexNeRF}, which have already been demonstrated with NeRFs. 
\rebuttal{Hence, object detection in NeRFs has the potential to benefit robot autonomy stacks based on NeRF environment representations.}
Our approach hopes to bridge this gap.

Specifically, we study \emph{3D object localization in neural fields}, as a step towards bridging the gap between robotic perception and planning/control in environments represented by NeRFs. 
The challenge mainly lies in how to effectively exploit the geometric information contained in the NeRF representation, and in particular, to take advantage of the ease of scaling with a NeRF representation.
Specifically, we design a transformer-based network to directly estimate object localization within a Neural Radiance Field. 
To take advantage of the ease of scaling up and down with the NeRF representation, our framework includes two sub-networks of coarse and fine streams, which efficiently fuse the information from the wider horizon and the zoom-in view for comprehensive scene understanding (Fig.~\ref{fig:teaser}).  
Intuitively, the coarse stream perceives a broader view which provides more global contexts to alleviate the ambiguities, while the fine stream helps to localize the object at a finer level. Our model takes in a previously unseen pre-trained NeRF model and camera view, and outputs labeled 3D bounding boxes for the objects in that NeRF.

Currently, there is a scarcity of datasets for our task, which needs to have both NeRF representations and 3D bounding box annotations. Objectron~\cite{objectron2021} is a dataset 
recently proposed suitable for our task. 
It contains consecutive video frames and corresponding camera poses, which can be used for NeRF training.
We build a NeRF object localization benchmark \benchmarkName based on Objectron~\cite{objectron2021} and evaluate our method with it. 
We show that our method outperforms previous baselines.% while being ?x 

In summary, our main contributions are as follows:
\begin{itemize}
    \item We introduce the problem of \emph{object localization in a neural radiance world}, a step towards semantic robotic perception with neural scene representations, which can be used for downstream tasks such as planning and control.
    \item We propose \algoName, a framework for 3D object localization that exploits the geometric information imposed by neural representations. 
    \item  We evaluate our approach extensively and experimental results show that our approach significantly outperforms existing methods in NeRF object localization task.
\end{itemize}

\begin{algorithm}[t]
	\caption{3D Object Localization on Neural Fields (Training Process)}
	\label{alg:algo_training}
	\KwData{NeRF function $F_\Theta$, camera pose $\mathbf{P}$, coarse intrinsics $K^C$ and fine intrinsics $K^F$, \rebuttal{ground truth $\psi=\{B, p\}$, where $B$ represents the bounding box instances and $p$ represents the classes.}}
    \KwResult{$J$ object proposals $\{\hat{\psi}^j\}_{j=1}^J$, $\hat{\psi}^j=\{\hat{B}^j, \hat{c}^j\}$, $\hat{B}=\{x_i^b, y_i^b, z_i^b\}_{i=1}^{N_c}$ is the coordinates for each bounding box. $N_c$ is the number of corner points. $\hat{c}$ is the predicted class for each object.}
	$\bm{X}^C = F_\Theta(\mathbf{P}, K^C)$;
	
	$\bm{X}^F = F_\Theta(\mathbf{P}, K^F)$;
	
	Compute embeddings using coarse encoder and fine encoder (see Equ.~\eqref{equ:coarse_stream} and Equ.~\eqref{equ:fine_stream});
	
	Fuse embeddings using attention fusion module (see Equ.~\eqref{equ:fusion_module});
	
	Compute output embeddings using decoder; % (see Equ.~\eqref{equ:decoder});
	
	\For {$j \in 1, 2, \cdots, J$}
	{
	Predict bounding box instance $\hat{\psi}^{j}=\{\hat{B}^j, \hat{p}^j\}$ using prediction head;
	}
	Compute optimal matching $\sigma*$ between $\{\psi^{j}\}^J_{j=1}$ and $\{\hat{\psi}^{j}\}^{J}_{j=1}$ using Hungarian matcher;
	
	Compute final loss $\mathcal{L}_H$ between $\{\psi^{j}\}^J_{j=1}$ and $\{\hat{\psi}^{\sigma*{(j)}}\}^J_{j=1}$;
	
	Backpropagate $\mathcal{L}_H$;
\end{algorithm}

\section{Related Work}
\noindent \textbf{Neural Radiance Field (NeRF).} NeRF~\cite{mildenhall2020nerf} utilizes an MLP network to predict the density and color of points in a scene, which allows for differentiable rendering by tracing rays through the scene and integrating them. 
Semantic NeRF~\cite{Zhi:etal:ICCV2021} extends NeRF to jointly encode 2D semantics with appearance and geometry. 
There are also some few-shot NeRFs~\cite{yu2020pixelnerf} to perform novel view synthesis from a sparse set of views. 
Recently, object-centric NeRFs investigate how the synthesis process can be controlled at the object level. GIRAFFE~\cite{Niemeyer2020GIRAFFE} incorporates a compositional 3D scene representation into the generative model which leads to more controllable image synthesis. STaR~\cite{yuan2021star} jointly optimizes the parameters of two Neural Radiance Fields and a set of rigid poses to decompose a dynamic scene into two constituent parts. Impressively, Block-NeRF~\cite{tancik2022blocknerf} demonstrates the possibility to scale NeRF to render city-scale scenes spanning multiple blocks.
With such great advances in existing NeRF-related technologies, 
researchers recently have explored representing the scene with NeRFs in robotic applications~\cite{adamkiewicz2022vision,li20213d,IchnowskiAvigal2021DexNeRF}. Following these works, we discuss object localization within the NeRF scenes in this paper, in the hope of facilitating downstream robotic applications with NeRF representations. 

\noindent \textbf{Object Localization.}
Object localization is a key component for robotics applications including autonomous driving, indoor navigation, and robot manipulation. 
Most previous object localization methods can be divided into three main categories according to the input modality types: point-cloud based, stereo images based, and monocular image based. 
The point-cloud based methods\rebuttal{~\cite{chen2017multi,qi2018frustum,shi2019pointrcnn,yi2020segvoxelnet,wang2019densefusion,he2020pvn3d,he2021ffb6d,di2022gpv,peng2022self}}  directly acquire the coordinates of the points on the surfaces of objects in 3D space. These methods generally work on the point clouds obtained from hardware Time of Flight (ToF) sensors.  
Despite good performance, the costly depth sensor is not always available for a robotic system. Stereo image based methods~\cite{chen20173d} leverage the geometric structures obtained from the disparities between the stereo image pair. 
The monocular methods~\cite{xiang2017posecnn,9009519,li2019gs3d,ding2019iccv} become popular for object localization in the community given the portable and low-cost nature.  
\rebuttal{However, our approach is specialized for NeRF-based scene representation.}

\noindent \textbf{Implicit Representations for Robotics.} 
Adamkiewicz~\etal propose a trajectory planning method that plans full, dynamically feasible trajectories to avoid collisions with a NeRF environment~\cite{adamkiewicz2022vision}. Li \etal~\cite{li20213d} combine NeRF and time contrastive learning to learn viewpoint-invariant 3D-aware scene representations, which enables visuomotor control for challenging manipulation tasks. Dex-NeRF~\cite{IchnowskiAvigal2021DexNeRF} leverages NeRF's view-independent learned density, and performs a transparency-aware depth-rendering to grasp transparent objects. Lin \etal~\cite{yen2022nerfsupervision} propose to learn dense object descriptors from NeRFs and use an optimized NeRF to extract dense correspondences between multiple views of an object. 
LENS~\cite{moreau2021lens}, \rebuttal{NeRF-Pose~\cite{li2022nerf}, Loc-NeRF~\cite{maggio2022loc}}, and iNeRF~\cite{yen2020inerf} proposed camera localization algorithms based on NeRF representation. Their ideas have the potential to be applied to object pose estimation but may be prone to errors given complex environments and multi-object cases. 
In contrast to previous work, we find the compact representation of NeRF (which can be zoomed in and out freely) is particularly useful for 3D object localization, which can further aid downstream robotics applications such as navigation and manipulation. In particular, unlike LENS~\cite{moreau2021lens}, our detection framework can do multi-object detection.
\begin{figure}
    \centering
    \includegraphics[width=0.9\linewidth]{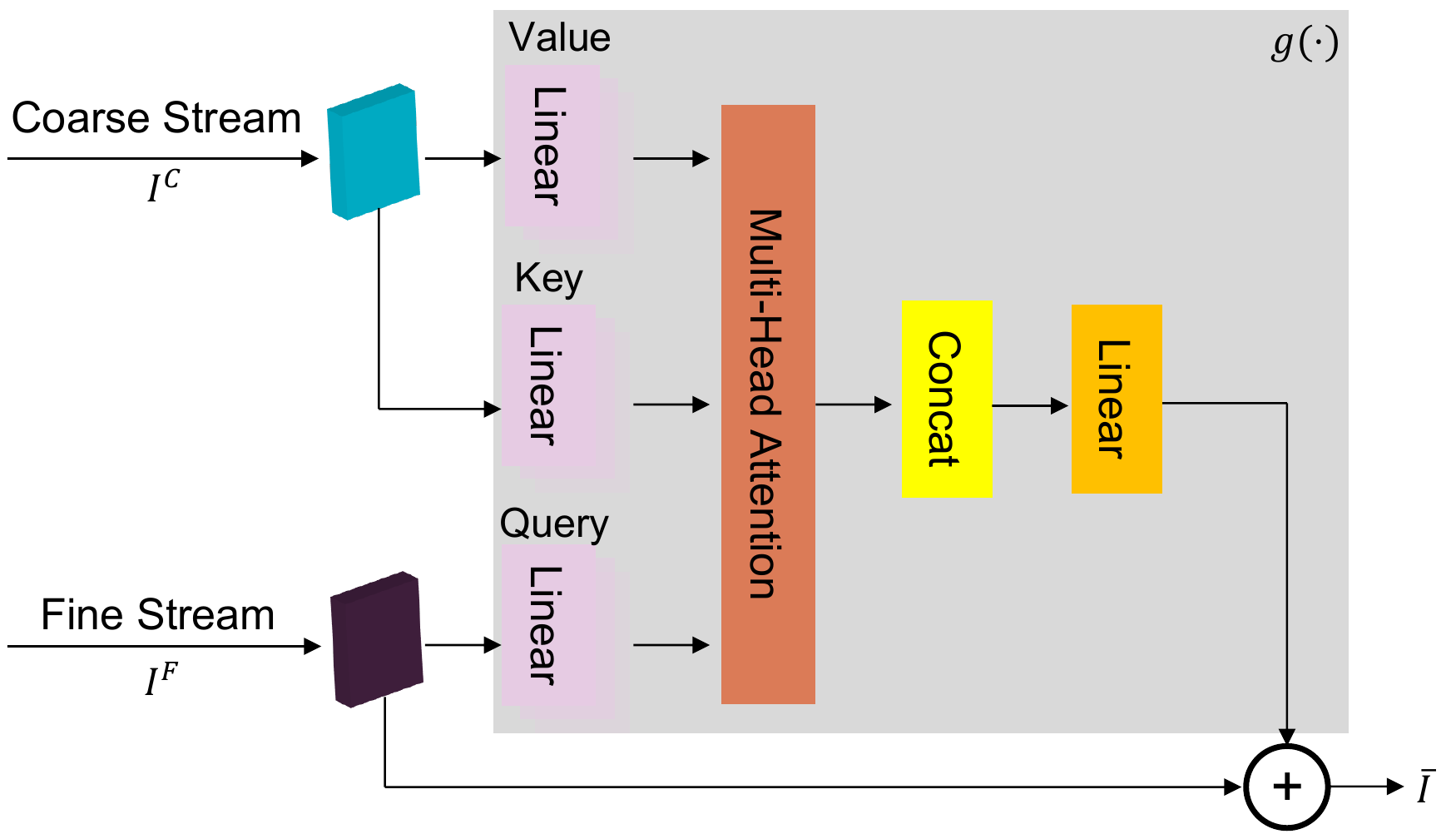}
    \caption{Cross-attention Fusion Module fuses abstraction levels of coarse and fine context. 
}
\label{fig:fusion_module}
\vspace{-10pt}
\end{figure}
\section{Method}
\subsection{Preliminary}
NeRF~\cite{mildenhall2020nerf} is a differentiable implicit function that represents a continuous 3D scene. 
The implicit function is usually implemented with Multi-Layer Perceptrons (MLPs) $F_\Theta : (\bm{x}, \bm{d}) \rightarrow (\bm{c}, \sigma)$, which maps the 3D location $\bm{x} = (x, y, z)$ and 2D viewing direction $\bm{d}=(\theta, \phi)$ to  an emitted color value $\bm{c}$ and a volume density value $\sigma$. 
Based on this representation, the pixel color can be obtained via volume rendering~\cite{levoy1990efficient}: 
\begin{equation}
\begin{split}
    \hat{C}(\bm{r}) = \sum_{k=1}^{K}T_k\left(1-\exp\left(-\sigma_k(t_{k+1}-t_k)\right)\right)\bm{c}_k, \\ 
    % \rm{where}\  T_k=\exp\left(-\sum_{k'<k}\sigma_{k'}(t_{k'+1}-t_{k'})\right), 
\end{split}
\end{equation}
where $T_k=\exp\left(-\sum_{k'<k}\sigma_{k'}(t_{k'+1}-t_{k'})\right), \bm{r}(t) = \bm{o} + t\bm{d}$ denotes a ray cast from the camera center $\bm{o}$ along the direction $\bm{d}$ passing through the rendering pixel.
$T_k$ here can be interpreted as the probability that the ray is not interrupted before and successfully transmits to point $\bm{r}(t_k)$.
Similarly, the expected depth $\hat{D}(\bm{r})$ where the camera ray $\bm{r}(t) = \bm{o} + t\bm{d}$ terminates can be calculated by replacing the color value $\bm{c}_k$ with the sampling distances $t_k$: 
\begin{equation}
    \begin{split}
    \hat{D}(\bm{r}) = \sum_{k=1}^{K}T_k\left(1-\exp\left(-\sigma_k(t_{k+1}-t_k)\right)\right)t_k. \\ 
\end{split}
\end{equation}

\subsection{Problem Formulation}
We define the \emph{object localization within Neural Radiance Fields} task as follows:
Given a pre-constructed NeRF environment $F_\Theta$ and an observation pose $\mathbf{P}\in SE(3)$ inside, we design a transformer network $G_\Phi$ to estimate the 3D bounding box $\hat{B}$ and category $\hat{p}$ of $J$ objects in the current view:
\begin{equation}
    \{(\hat{B}^j, \hat{p}^j)\}_{j=1}^J = G_\Phi(\mathbf{P}; F_\Theta ). 
    \label{equ:nerfdet_1}
\end{equation}
Here, the bounding box $\hat{B}$ is parameterized as its corners $(\hat{x}^b, \hat{y}^b, \hat{z}^b)$:  
$\hat{B}=\{(\hat{x}^b_i, \hat{y}^b_i, \hat{z}^b_i)\}_{i=1}^{N_c}$ ($N_c=8$ in our case). 

\begin{figure}
    \centering
    \subfloat[]{\includegraphics[width=0.3\linewidth]{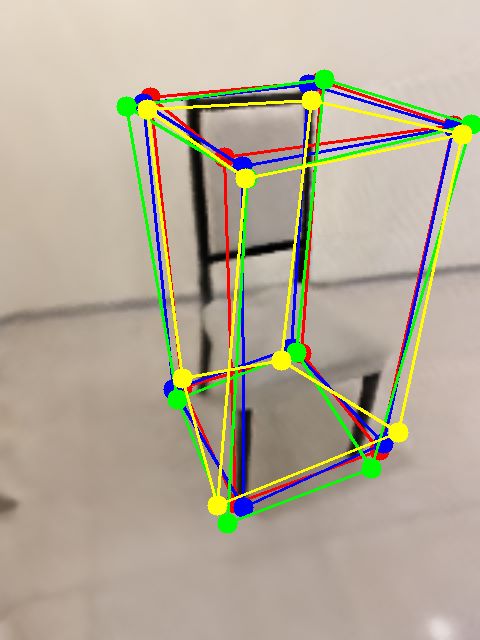}\label{}}
    \hspace{0.01cm}
    \subfloat[]{\includegraphics[width=0.3\linewidth]{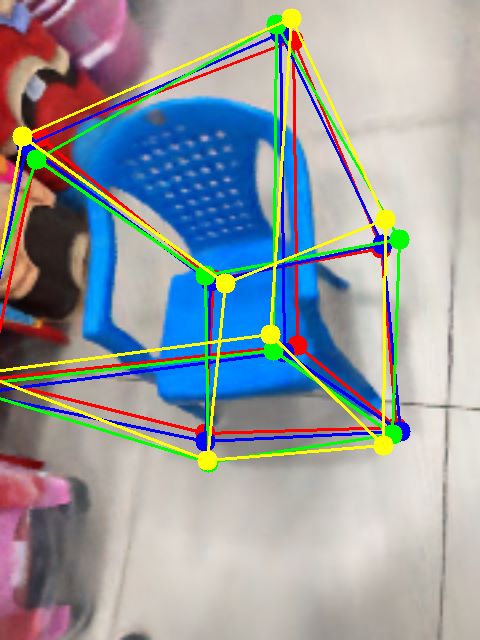}\label{}}
    \vspace{0.1cm}
    \subfloat[]{\includegraphics[width=0.3\linewidth]{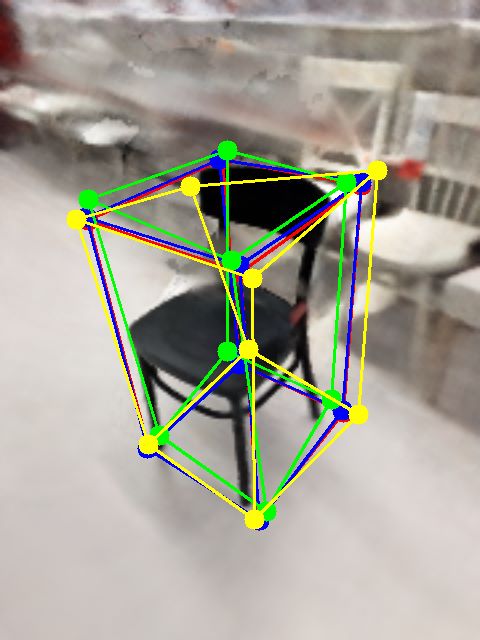}\label{}}
    \vspace{-0.1cm}
    \caption{
    Qualitative results of \algoName on the validation split. The detected objects are shown with 3D bounding boxes. Groundtruths are labeled with \textcolor{red}{red} while the predictions from NeRF samples $\hat{\bm{X}}$ are labeled with \textcolor{blue}{blue}, the predictions from rendered color images $\hat{C}(\bm{r})$ are labeled with \textcolor{green}{green}, the predictions from rendered depth maps $\hat{D}(\bm{r})$ are labeled with \textcolor{yellow}{yellow}.}
    \label{fig:qual_results}
    \vspace{-10pt}
\end{figure}
\vspace{-10pt}
\subsection{\algoName}
Inspired by the effectiveness of multi-view information~\cite{pan2020cross} and the outstanding performance of transformer-based methods in localization~\cite{misra2021-3detr,sun2022locate}, we designed a transformer-based framework to efficiently utilize information from multiple views.
Fig.~\ref{fig:framework} shows an overview of the proposed framework. 
The pipeline consists of the following steps: 1) 
Given an observation pose $\mathbf{P}\in SE(3)$ and the camera intrinsic matrix $\mathbf{K}$, the field values (i.e. colors and densities) on the \rebuttal{rays} emitted from the camera center are sampled;  
2) 
The sampled field values are then sent to a transformer-based coarse encoder and fine encoder for feature extraction. 
3) These encoded features are thereafter fused with an attention fusion module to complement each other and alleviate ambiguity. 
4) Finally, the fused features are sent to the transformer-based decoder to predict the bounding-box corners and categories.

\subsubsection{Fine Stream and Coarse Stream}
To localize an object in the scene, the network not only should focus on the object itself but also should leverage the helpful context information around \rebuttal{e.g., scenes, or information around objects}. Following this intuition, we design two parallel branches, the fine stream and the coarse stream, to focus on the object details and the context respectively.

Given an observation pose $\mathbf{P}\in SE(3)$ and the camera intrinsics $\mathbf{K} =\begin{bmatrix} f &0 &p_x \\ 0 &f &p_y\\ 0 &0 &1 \end{bmatrix}$,  the rays from the camera center passing through the image plane are chosen for sampling the radiance field. 
Specifically, each ray direction $\bm{d}$ is related to the focal length $f$ and the intersection points $(x,y)$ on the image plane: 
\begin{equation}\label{eq:ray_direction}
\bm{d}(x,y,f) = \frac{[x-p_x, y-p_y, f]^T}{\sqrt{(x-p_x)^2+(y-p_y)^2+ f^2}}. 
\end{equation}

We apply camera intrinsic matrices
with two different focal lengths, $f/\delta$ and $f$,  where $\delta>1$, for the coarse stream and the fine stream respectively.  
In our case, we set $\delta = 1.5$. 
In this way, different sampling scopes are adopted for different streams according to Equ.~\eqref{eq:ray_direction}. The coarse stream essentially has a larger field of view while the fine stream has more detailed views. Empirically, we find the two-stream encoding significantly improves the localization performance. 

After having the sampling rays mentioned above, we sample $N$ equally-spaced points on each ray and collect the field values $(\mathbf{c}, \sigma)$ on these points. 
The sampled coarse point sets $\hat{X}^C$ and fine point sets $\hat{X}^F$ are modulated by projection layers $l^C(\cdot)$ and $l^F(\cdot)$ respectively, 
and then sent to the coarse and fine transformer encoders $E^C(\cdot)$ and $E^F(\cdot)$ for further processing:
\begin{equation}
\label{equ:coarse_stream}
    I^F = E^F(l^F(\hat{X}^F)),
\end{equation}
\begin{equation}
\label{equ:fine_stream}
    I^C = E^C(l^C(\hat{X}^C)).
\end{equation}
After the processing, the coarse embeddings $I^C$ and fine embeddings $I^F$ are obtained. 

\begin{figure}
    \centering
    \subfloat[]{\includegraphics[width=0.3\linewidth]{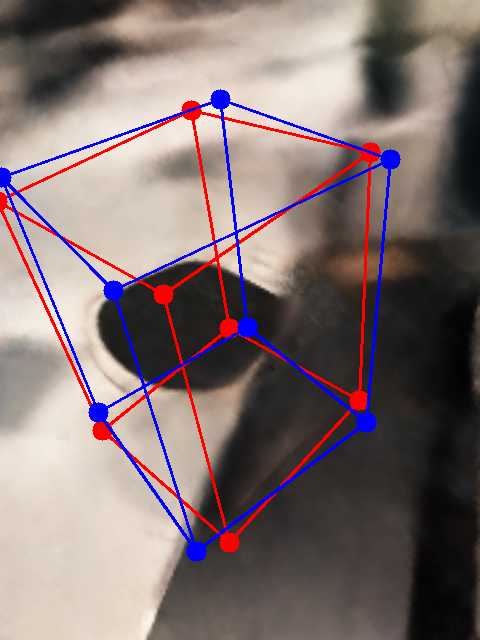}\label{}}
    \hspace{0.01cm}
    \subfloat[]{\includegraphics[width=0.3\linewidth]{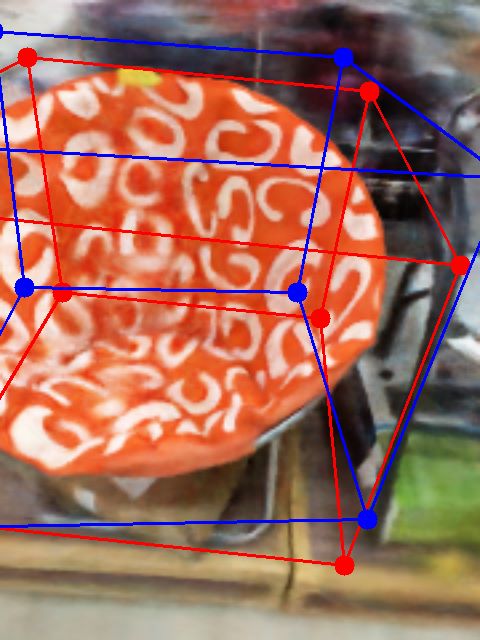}\label{}}
    \vspace{0.1cm}
    \subfloat[]{\includegraphics[width=0.3\linewidth]{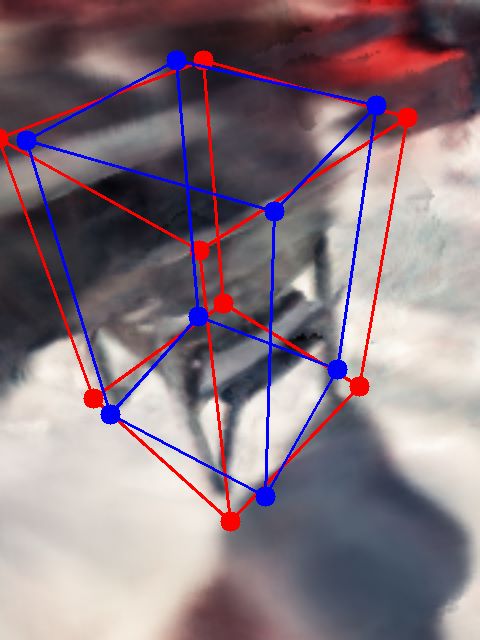}\label{}}
    \vspace{-0.1cm}
    \caption{
    Failure cases.
    a) Prediction error is caused by severely blurred rendering. b) The object occlusion leads to a failure case. c) The color similarity between the foreground and background poses difficulties for object localization.}
    \label{fig:failure_case}
    \vspace{-10pt}
\end{figure}
\subsubsection{Cross-attention Fusion}
To make better use of wide-view information and fine-grained information, we introduce Cross-attention Fusion, a lateral connection between the coarse stream and fine stream, to fuse the embeddings from the fine stream and the coarse stream (Fig.~\ref{fig:fusion_module}). 
The fine stream embeddings $I^F$ and coarse stream embeddings $I^C$ are calculated through a lightweight cross-attention head $g(\cdot)$. Then, the selected information is fused with the original fine-grained information via skip connection as Equ.~\eqref{equ:fusion_module} shows. We hope that such a design could learn and benefit from both fine-grained contexts and coarse-grained contexts at the abstraction level.
\begin{equation}
\label{equ:fusion_module}
    \bar{I} = I^F + g(I^F, I^C).
\end{equation} 

\subsubsection{Decoder}
The fusion embeddings $\bar{I}$ are passed through the transformer decoder network $D(\cdot)$, and $J$ localization proposals are obtained from MLP heads, as expressed by Equ.~\eqref{equ:decoder}.
\begin{equation}
\label{equ:decoder}
    \{(\hat{B}^j, \hat{p}^j)\}_{j=1}^J = D(\bar{I}, q),
\end{equation}
$q$ is the learnable query in the shape of $J$ which is randomly initialized. \rebuttal{
Please refer to ~\cite{carion2020end} for an explanation of the usage of $q$, which is a commonly used design in transformers.} 

\subsection{Loss Functions}
From the embeddings output from the decoder, 3D bounding box corners are predicted by MLP regression head $\hat{B}$, and the corresponding categories $\hat{p}$ are predicted by the MLP classification head.
The optimal match is computed between the augmented ground-truth $\psi^{(j)}$ and prediction $\hat{\psi}^{(j)}$. We search for the optimal permutation $\sigma*$ among the set of all permutations, that has the lowest matching cost $\mathcal{L}_{match}$. 
\begin{equation}
    \sigma*={\arg\min}_{\sigma\in\sum_{J}}\sum_{j=1}^{J}\mathcal{L}_{match}(\psi^j, \hat{\psi}^{\sigma(j)}).
\end{equation}
This cost is computed efficiently via the Hungarian algorithm. $\mathcal{L}_{box}$ is a weighted combination between the IoU loss $\mathcal{L}_{iou}$~\cite{union2019metric} and $\ell_1$ loss, weighted by scalar hyperparameters $\lambda_{iou}$ and $\lambda_{\ell_1}$
\begin{equation}
    \label{equ:loss_obj}
    % \mathcal{L}_r=\sum_{j\in J}\sum_{i\in(1,\dots, N_c)}\left(||x_i-\hat{x}_i||_2 + ||y_i-\hat{y}_i||_2 + ||z_i-\hat{z}_i||_2\right)
    \mathcal{L}_{box}=\lambda_{iou}\mathcal{L}_{iou}(B^j, \hat{B}^{\sigma*(j)}) + \lambda_{\ell_1}\ell_1(B^j, \hat{B}^{\sigma*(j)}).
\end{equation}
We use cross-entropy loss $\mathcal{L}_{CE}$ as classification objective.
The Hungarian matching loss $\mathcal{L}_{match}$ is a sum of the classification and regression loss $\mathcal{L}_{match}(\psi^n, \hat{\psi}^{\sigma*(n)}) = \mathcal{L}_{box} + \mathcal{L}_{CE}$.
After obtaining the optimal permutation $\sigma*$ based on the lowest $\mathcal{L}_{match}$, we compute the Hungarian loss $\mathcal{L}_H$ for this optimal matching. $\mathcal{L}_H$ over all the matched pairs of proposals is defined as:
\begin{equation}
\label{equ:loss_total}
    \mathcal{L}_H=\sum_{j=1}^{J}\mathcal{L}_{match}(\psi^j, \hat{\psi}^{\sigma*(j)}).
\end{equation}
\algoName is trained end-to-end, with $\mathcal{L}_H$ as its objective. The full \algoName learning pipeline is illustrated as Alg.~\ref{alg:algo_training}.

\begin{table}[t]
\caption{Performance comparison of our approach and baselines. For single-view, the fine view is used for baselines. For multi-view, both fine and coarse information is used for baselines.}
\vspace{-6pt}
\begin{center}
\resizebox{\linewidth}{!}{
\begin{tabular}{c|c|c|c|c|c}
\hline
\multirow{2}{*}{\textbf{\shortstack{Algorithm\\ Category}}} & \multirow{2}{*}{\textbf{Method}} &\multicolumn{3}{|c|}{\textbf{mAP@IoU (\%)}} & \multirow{2}*{Average} \\
\cline{3-5} 
& & 0.1 & 0.5 & 0.9 &  \\
\hline
\multirow{4}{*}{Single-view} & \texttt{3DETR~\cite{misra2021-3detr}} & 78.15 & 56.96 & 0.11 & 50.20  \\
& \texttt{CDPN~\cite{9009519}} & 95.01 & 83.39 & 0.49 & 66.28 \\
& \algoName (Fine-only) & 98.24 & $\bm{91.49}$ & 0.55 & 70.78 \\
& \algoName (Coarse-only) & 95.50 & 88.29 & 0.25 & 67.67 \\
\midrule
\multirow{3}{*}{Multi-view} & \texttt{3DETR~\cite{misra2021-3detr}} & 80.21 & 59.12  & 0.23 & 52.25  \\
& \texttt{DETR3D~\cite{wang2022detr3d}} & 97.42 & 86.28 & 0.21 &  67.84 \\
 & \algoName & $\bm{99.22}$ & 87.92 & $\bm{1.70}$ & $\bm{72.02}$  \\
\hline
\end{tabular}}
\label{tab:performance}
\end{center}
\vspace{-10pt}
\end{table}
\section{Experimental Results}

We aim to answer the following questions in our experiments: (i) Can we learn to predict object bounding boxes in Neural Fields? How does \algoName
compare to existing methods? 
(ii) Is the raw representation of NeRF better than the other representations?

\subsection{Dataset} 
To train and test our proposed model, we build a benchmark \benchmarkName based on Objectron~\cite{objectron2021}.
Each video has longer clips averaging a duration of $\sim$15 seconds. Such a long duration makes it a suitable
dataset to train NeRF models and test \algoName networks. 
The \benchmarkName consists of coarse and fine-grained NeRF samples $\hat{X}$ trained using NeRF~\cite{mildenhall2020nerf}, rendered color image $\hat{C}(\bm{r})$, rendered depth image $\hat{D}(\bm{r})$ and corresponding 3D bounding box annotations. 

\subsection{Baselines}
Many previous 3D object localization methods rely on CAD models, which are not required by us. We evaluate the capacity of our model in 3D object localization in neural fields and compare it with the multiple baselines and ablations:
Among them, 3DETR~\cite{misra2021-3detr} is a transformer-based object detection model. 
CDPN~\cite{9009519} estimates 6-DoF object pose estimation from RGB images. DETR3D~\cite{wang2022detr3d} performs 3D object detection from multi-view images. 

To enable fair comparison, we modify the above baselines to take the NeRF samples as input and predict the 3D bounding boxes as output. 
\subsection{Implementation Details}
We initialize the Coarse and Fine streams of our network from scratch and follow an end-to-end training process. 
We use $\delta=1.5, N_c=8, J=100$ in our experiments. The number of layers of transformer encoder and decoder $L^F_e = L^C_e = L_d=4$. For the feature projection layer $l(\cdot)$, we use 3 layers of MLPs to map the input to 256-dim. Both streams are trained together for 500 epochs with a batch size of $8$ and a cosine dynamic learning rate scheduler of $1e-6$ at the start, $5e-4$ as the base learning rate, and $9$ warm-up epochs with a warm-up learning rate of $1e-6$.
\rebuttal{$240\times180$ directions are sampled and 64 points along a pixel ray are used for forming an $X$ for both training and inference.}
We train our model with a machine using NVIDIA TITAN Xp GPU and Intel Xeon CPU.

\begin{figure}
    \centering
    \subfloat[]{\includegraphics[width=0.24\linewidth]{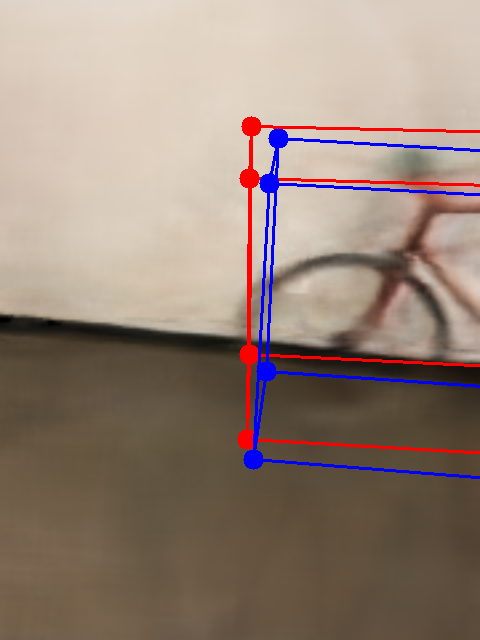}\label{}}
    \hspace{0.01cm}
    \subfloat[]{\includegraphics[width=0.24\linewidth]{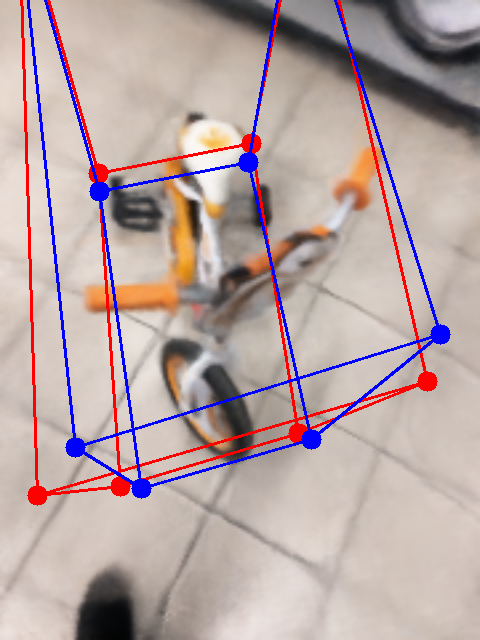}\label{}}
    \vspace{0.1cm}
    \subfloat[]{\includegraphics[width=0.24\linewidth]{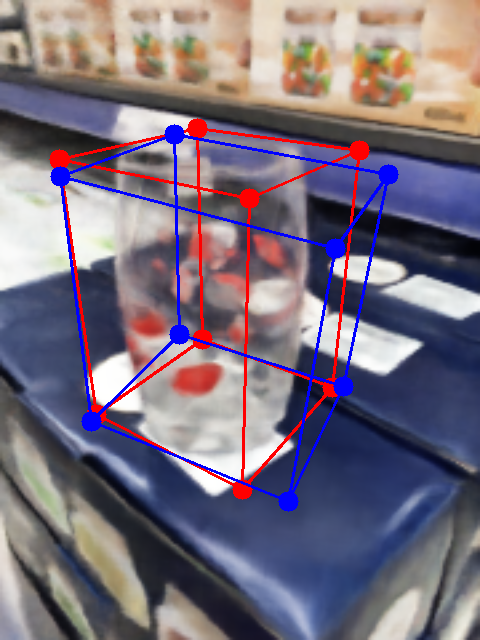}\label{}}
    \vspace{-0.1cm}
    \subfloat[]{\includegraphics[width=0.24\linewidth]{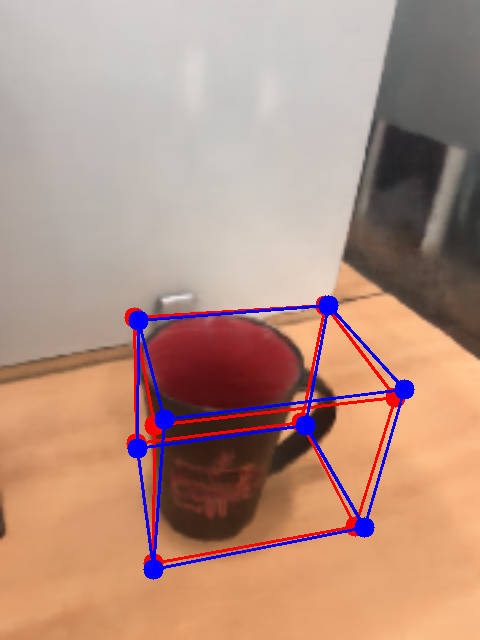}\label{}}
    \caption{Qualitative results of unseen scenes and novel views. 
    \rebuttal{(a) (b) are unseen scenes and (c) (d) are unseen views.}}
    \label{fig:novel_view_scene}
 \vspace{-2ex}
\end{figure}

\subsection{Evaluating \algoName} % on \benchmarkName}
First, we evaluate \algoName on \benchmarkName by only taking NeRF samples as input to verify that we can predict object bounding boxes in neural fields.

In Fig.~\ref{fig:qual_results}, Fig.~\ref{fig:failure_case},  Fig.~\ref{fig:novel_view_scene}, we visualize some examples of the predicted bounding boxes results on \benchmarkName. Our model is able to predict the bounding box of objects with the correct position. 
We compare the performance of the \algoName with recent methods on the 3D object localization task in Tab.~\ref{tab:performance}.
For this evaluation, we report the performance (mAP) at different IoUs. 
The comparisons in Tab.~\ref{tab:performance} show that our approach performs better than the previous methods and multi-view input has higher performance than single-view input. Our approach is more effective than other methods when the mAP is at high IoU. 
\rebuttal{Sometimes NeRF is not trained well enough (e.g., it is rendered blurry in Fig.~\ref{fig:failure_case}), which can lead to poor detection. Our choice of coarse- and fine-grained scale (e.g., the field of view does not contain sufficient information about the object) can also affect performance.}
Although our model outperforms the multi-view baselines using only the fine view, the proposed Coarse-Fine architecture further improves it by a large margin (from 70.78\% to 72.72\%).
\begin{table}[]
\caption{Ablation study of different modalities. Both fine and coarse information is used. $\hat{D}(\bm{r})$: rendered depth image, $\hat{C}(\bm{r})$: rendered color image, $\hat{\bm{X}}$: NeRF raw samples. }
\vspace{-6pt}
\setlength\tabcolsep{7pt}
\begin{center}
\resizebox{0.8\linewidth}{!}{
\begin{tabular}{c|c|c|c|c}
\hline
\textbf{Modality}&\multicolumn{3}{|c|}{\textbf{mAP@ (\%)}} & \multirow{2}*{Average} \\
\cline{2-4} 
\textbf{IoU} & 0.1 & 0.5 & 0.9 &  \\
\hline
$\hat{C}(\bm{r})$ & 81.92 & 49.07 & 0.17 & 43.01  \\
$\hat{D}(\bm{r})$ & 78.51 & 33.74 & 0.00 & 35.28  \\
$\hat{\bm{X}}$ & 99.22 & 87.92 & 1.70 & 72.02 \\
$\hat{\bm{X}}$ + $\hat{C}(\bm{r})$ & 98.29 & 93.11 & 0.46 & $\bm{72.78}$  \\
$\hat{\bm{X}}$ + $\hat{D}(\bm{r})$ & 97.81 & 84.97 & 1.05 & 68.43  \\
$\hat{C}(\bm{r})$ + $\hat{D}(\bm{r})$ & $\bm{99.34}$ & $\bm{94.02}$ & 0.00 & 67.92  \\
$\hat{\bm{X}}$ + $\hat{C}(\bm{r})$ + $\hat{D}(\bm{r})$ & 97.91 & 86.05 & $\bm{1.84}$ & 69.80  \\
\hline
\end{tabular}}
\label{tab:abl_study_modality}
\end{center}
\vspace{-10pt}
\end{table}

\begin{table}[]
\caption{Ablation study of different fusion modules.}
\vspace{-6pt}
\begin{center}
\setlength\tabcolsep{8pt}
\resizebox{0.8\linewidth}{!}{
\begin{tabular}{c|c|c|c|c}
\hline
\textbf{Method}&\multicolumn{3}{|c|}{\textbf{mAP@ (\%)}} & \multirow{2}*{Average} \\
\cline{2-4} 
\textbf{IoU} & 0.1 & 0.5 & 0.9 &  \\
\hline
MLP & 98.95 & $\bm{88.08}$ & 1.45 & 67.60  \\
Attention & $\bm{99.22}$ & 87.92 & $\bm{1.70}$ & $\bm{72.02}$ \\
\hline
\end{tabular}}
\label{tab:abl_study_fusion_type}
\end{center}
\vspace{-10pt}
 \vspace{-2ex}
\end{table}

\subsection{Tracking with NeRF Representation}
We show the qualitative results of pose tracking in Fig.~\ref{fig:tracking}. Once we got the pre-trained NeRF, at each time step $t$, \algoName estimates the object’s bounding box, without the need to remove the background. 
\begin{table}[t]
\begin{center}
\renewcommand\tabcolsep{2pt}
\begin{small}
% \begin{sc}
\begin{tabular}{>{\centering\arraybackslash}m{0.22\linewidth}>{\centering\arraybackslash}m{0.22\linewidth}>{\centering\arraybackslash}m{0.22\linewidth}>{\centering\arraybackslash}m{0.22\linewidth}
}
\toprule
$t=1$ & $t=2$  & $t=3$  & $t=4$ \\
\midrule
\includegraphics[width=\linewidth]{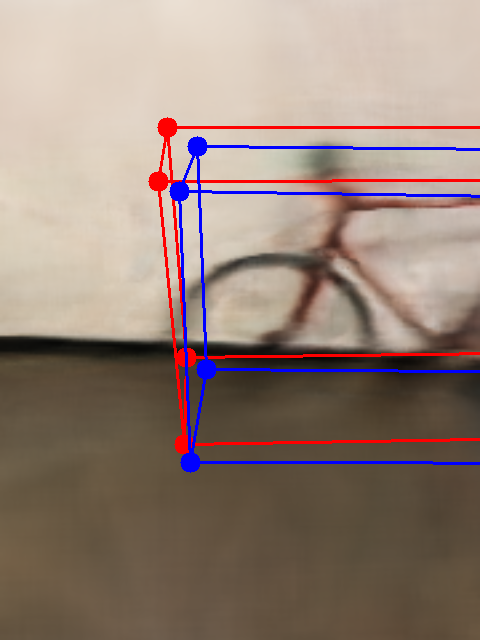} & \includegraphics[width=\linewidth]{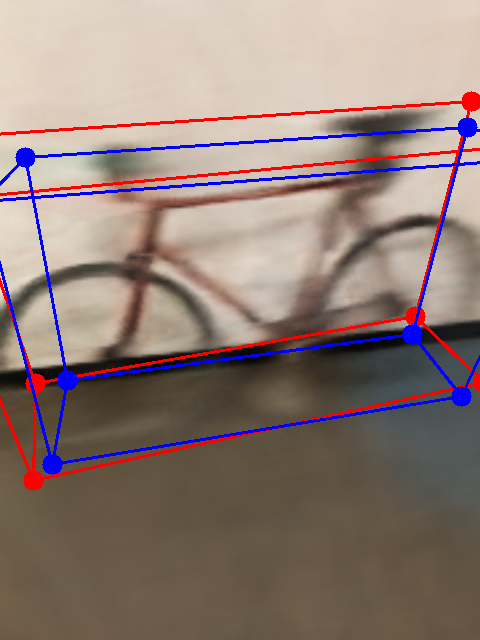}  & \includegraphics[width=\linewidth]{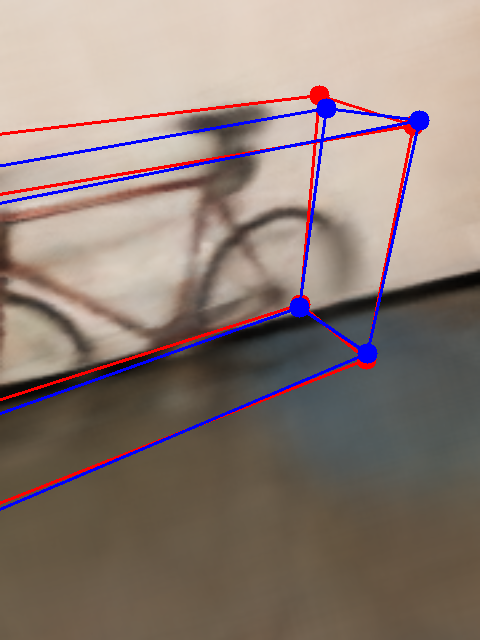} & \includegraphics[width=\linewidth]{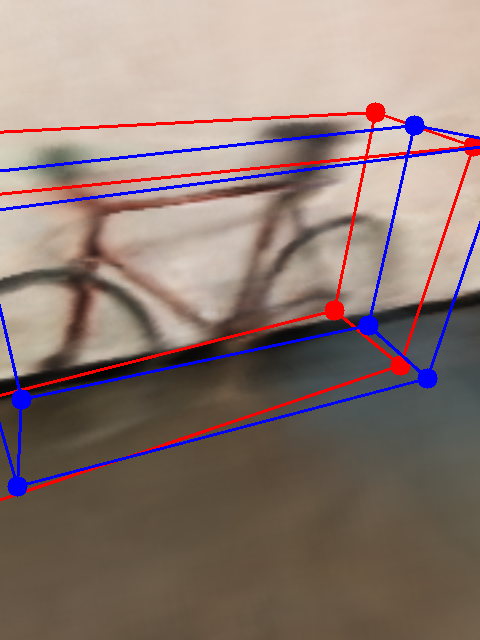} \\
\bottomrule
\end{tabular}
\end{small}
\captionof{figure}{Qualitative results of object tracking in NeRF representations without the need for removing the background.}
\label{fig:tracking}
\end{center}
\vspace{-10pt}
\end{table}

\subsection{Ablation Study}
\subsubsection{Modality}
We compare multiple modality combinations: (i) NeRF samples $\hat{\bm{X}}$, (ii) rendered color image $\hat{C}(\bm{r})$, and (iii) rendered depth image $\hat{D}(\bm{r})$, and the permutation of these three modalities. Both fine and coarse views are used for all modality combinations. We find that using only the NeRF samples is better than using the rendered color image or the rendered depth image individually
(see Tab.~\ref{tab:abl_study_modality}). This indicates that NeRF samples are more effective than the other two representations and facilitate faster and better localization. The average mAP of $\hat{\bm{X}} + \hat{C}(\bm{r})$ is slightly better than using $\hat{\bm{X}}$ only. Considering the time spent on rendering, the obtained NeRF samples are sufficient for downstream tasks. There is a limited necessity to spend further time on rendering color or depth maps.

\subsubsection{Fusion Type} We have also tried MLP fusion layers: simply passing $I^C$ and $I^F$ through 2 layers of MLPs after stitching them in the last dimension to get $\bar{I}$.
As Tab.~\ref{tab:abl_study_fusion_type} shows, we can see that our cross-attention fusion module works much better than the MLP fusion module. The difference is more significant especially when the IoU threshold is high. This is probably due to the fact that the attention mechanism is more sensitive to fine-grained features.

\section{Discussion, Limitations, and Conclusions
}
We propose the task of object localization in a neural radiance world, which enables autonomous agents to perceive under implicit representation, understand where the goal is, and used it for downstream tasks such as planning and control. We presented \algoName, a framework for 3D object localization in neural fields, which can take advantage of the ease of scaling up and down with the NeRF representation. We introduced the Cross-attention Fusion to best combine the coarse stream with the fine stream.  
\rebuttal{One of the limitations of our work is that we assume that the NeRF model is already pre-trained. The NeRF
model usually takes time for training and has difficulties being directly applied to real-time systems. With the progress of technology, the training time of NeRF has been greatly reduced.} Our work can be seen as a step towards the goal of enabling autonomous agents to find the target location from NeRF world and plan for complex tasks like navigation and manipulation. In future work, we intend to further explore object localization tasks in more complex NeRF scenarios.

\section*{Acknowledgment}
This work was partially supported by ONR grant N00014-18-1-2830.
\printbibliography
\end{document}